\documentclass[
]{ceurart}

\begin{document}

\copyrightyear{2022}
\copyrightclause{Copyright for this paper by its authors.
  Use permitted under Creative Commons License Attribution 4.0 International (CC BY 4.0).}

\conference{CERIST NLP Challenge 2022}

\title{Transformers and Ensemble methods: A solution for Hate Speech Detection in Arabic languages}

\author[1]{Angel Felipe Magnossão de Paula}[%
orcid=0000-0001-8575-5012,
email=adepau@doctor.upv.es,
]

\author[2,3]{Imene Bensalem}[%
orcid=0000-0002-2462-5967,
email=ibensalem@escf-constantine.dz,
]

\author[1]{Paolo Rosso}[%
orcid=0000-0002-8922-1242,
email=prosso@dsic.upv.es,
]

\author[4]{Wajdi Zaghouani}[%
orcid=0000-0003-1521-5568,
email=wzaghouani@hbku.edu.qa,
]

\address[1]{Universitat Politècnica de València}
\address[2]{MISC Lab – Constantine 2 University}
\address[3]{ESCF de Constantine}
\address[4]{Hamad Bin Khalifa University}

\begin{abstract}
This paper describes our participation in the shared task of hate speech detection, which is one of the subtasks of the CERIST NLP Challenge 2022. Our experiments evaluate the performance of six transformer models and their combination using 2 ensemble approaches. The best results on the training set, in a five-fold cross validation scenario, were obtained by using the ensemble approach based on the majority vote. The evaluation of this approach on the test set resulted in an F1-score of 0.60 and Accuracy of 0.86.
\end{abstract}


\maketitle

\section{Introduction}

The improvement of automatic hate speech detection is an important factor in diminishing the spread of toxicity online \cite{depaula2021detoxis}. Despite the recent advances in employing attention mechanisms and other deep learning approaches \cite{info13060273}, the detection of hate speech is still considered a major challenge \cite{magnossao-de-paula-etal-2022-upv}, especially, when dealing with social media text written in low-resource languages, such as Arabic and its various dialects. In fact, most of the naturally occurring Arabic text in social media is written in Dialectal Arabic (DA). 

The purpose of this paper is to present our approach to address hate speech detection in Arabic text. To this end, in addition to exploring six transformer-based architectures \cite{vaswani2017attention, devlin2018bert}, two ensemble methods are studied \cite{magnossao-de-paula-etal-2022-upv, depaula2021exist}. It should be noted that some of these architectures were specifically pre-trained in Arabic. Our code is open source  and available on GitHub.~\footnote{\url{https://github.com/AngelFelipeMP/Arabic-Hate-Speech-Covid-19}}

To carry out our experiments, we used data shared by the organizers of the CERIST NLP Challenge 2022 for task 1.d named Arabic hate speech and offensive language detection on social networks (COVID-19). The task is a binary classification problem where a model has to classify an Arabic tweet as Hateful or Not Hateful. The official evaluation metric for task 1.d is F1-score on the positive class (Hateful). In our experiments on the training set, the two highest F1-scores have obtained by employing the Majority Vote ensemble and AraBERT, respectively. Therefore, the Majority Vote ensemble is the method we have applied on the test set.

The remainder of the paper is structured as follows. Section 2 provides an overview of the problem of hate speech in Arabic. Sections 3 and 4 present the dataset details and the models applied. Finally, we close our paper by discussing the results and drawing some conclusions.

\section{Related Works}
In the past few years, the number of publications on hate speech detection in the Arabic language has taken a leap \cite{Husain2021}, especially with the organization of shared tasks addressing this research problem.

The first shared task has been organized within the 4th Workshop on Open-Source Arabic Corpora and Processing Tools (OSACT 4) \cite{Mubarak2020}. It addressed two binary classification tasks: offensive language detection and hate speech detection. The organizers provided a dataset of 10k tweets, wherein 20\% are offensive language and 5\% are hate speech. The best approach was obtained with the Support Vector Machine (SVM) model using an extensive pre-processing \cite{Hassan2020}. Another version of OSACT 4 dataset was also used in OffensEval shared task on multilingual offensive language identification \cite{Zampieri2020}. Besides the Arabic dataset, the organizers made available, to the participants, datasets in Danish, English, Greek, and Turkish. 

In addition to the binary classification tasks, the shared task organized within OSACT 5 Workshop \cite{Mubarak2022OSACT5} addressed the fine-grained hate speech categorization, where each hateful tweet has to be classified into one of the six following categories: race, religion, ideology, disability, social class and gender. The dataset (which was described in \cite{Mubarak2022Emoj}) is composed of more than 12k tweets, with 11\% labelled as hate speech. The top-ranked approach \cite{BenNessir2022} used a multitask model based on MARBERT and QRNN.

The detection of hatred against women (i.e., misogyny) has been addressed in ArMI shared task \cite{Mulki2021}, which proposed both binary and fine-grained classification subtasks. The dataset is composed of more than 9k tweets, where 61\% are misogynistic and labelled with one of 7 categories of misogyny.  The best system in this shared task \cite{Mahdaouy2021} combined the outputs of three different versions of MARBERT model using an ensemble approach.

The following sections are devoted to the description of the dataset and the experiments we conducted as part of our participation in the shared task of hate speech detection organized within the CERIST NLP challenge.

\section{Dataset}
The dataset of the hate speech detection task, shared by the organizers of the CERIST NLP challenge,  was collected from Twitter and split into training (80\%) and test (20\%) subsets. It consists of 10828 tweets, 11\% of which are annotated as hate speech. The domain of the dataset is COVID-19 disinformation. It is a multi-label dataset, which has been annotated not only for hate speech detection but also to tackle other tasks such as fake news detection. Further information on the dataset could be found in \cite{HadjAmeur2021}.

\section{Method}

This section presents the transformer models we applied to detect the Arabic hate speech and offensive language in social media  (COVID-19) for the challenge of task 1 proposed by the CERIST NLP Challenge 2022 organizers. The main features of our proposed transformed-based models are displayed in Table \ref{tab:transformers}.

\begin{table*}[!htp]
    \caption{Applied transformers to task 1.d}
    \label{tab:transformers}
    \begin{tabular}{lccccccc}\toprule
        \textbf{Version} & &\textbf{Size} & &\textbf{Block} & &\textbf{Language} \\\midrule
        AraBERT & &\multirow{2}{*}{Base} & &\multirow{3}{*}{Encoder} & &\multirow{4}{*}{Arabic} \\
        AraELECTRA & & & & & & \\\cmidrule(l){1-3} 
        Albert-Arabic & &Large  & & & \\\cmidrule(l){1-5} 
        AraGPT2 & &Base  & &Decoder & & \\\cmidrule(l){1-7}
        mBERT & &\multirow{2}{*}{Base} & &\multirow{2}{*}{Encoder} & &\multirow{2}{*}{Multilingual} \\
        XLM-RoBERTa & & & & & & \\
    \bottomrule
\end{tabular}
\end{table*}

A transformer is a massive deep learning model based on the self-attention mechanism \cite{vaswani2017attention,lin2021survey}. These models were originally built to handle natural language processing tasks \cite{ravichandiran2021getting}. The self-attention mechanism enables the transformer to focus on the crucial information from the input data, helping the model to achieve impressive results. Different unsupervised tasks are applied during the training process, such as mask language modelling, next sequence prediction, etc \cite{devlin2018bert,mohammed2021survey}. However, these models require large amounts of data to be trained.

Fortunately, some pre-trained transformers are freely available, and generally,  the users can select among three possible model sizes which are related to their number of trainable parameters: (i) Base, (ii) Medium, and (iii) Large. In the Table \ref{tab:transformers}’s second column, we can observe that apart from Albert-Arabic \cite{ali_safaya_2020_4718724} which we could use the Large pre-trained size, we adopted the smaller option (Base) for the models given our computational constraints regarding GPU memory. Albert-Arabic uses parameter reduction techniques to reduce the amount of memory required to allocate the pre-trained model to the GPU, which enables us to use its Large version.


The transformer's early architecture \cite{vaswani2017attention} was established based on an encoder and a decoder block. Nevertheless, the modern versions embody just one of those. As shown in Table  \ref{tab:transformers}’s third column, to solve task 1.d, we adopt five transformers based on the encoder block \cite{antoun2020arabert, antoun-etal-2021-araelectra, conneau2019unsupervised, ali_safaya_2020_4718724} and one transformer based on the decoder block \cite{antoun-etal-2021-aragpt2}.

Regarding the language of the text used for training step, the transformers can be divided into monolingual and multilingual models. The first is trained with monolingual data, which means text data in only one language (e.g., Arabic). The latter is trained with data in more than one language. We used four monolingual models trained in Arabic: AraBERT \cite{antoun2020arabert},  AraELECTRA \cite{antoun-etal-2021-araelectra},  Albert-Arabic, and AraGPT2 \cite{antoun-etal-2021-aragpt2}. Furthermore, we employed two multilingual models trained with documents in around 100 languages: mBERT \cite{devlin2018bert} and XLM-RoBERTa \cite{conneau2019unsupervised}.

\section{Results and Discussion}

This section describes the transformers’ hyper-parameter selection and the five-fold cross-validation carried out during the training phase. Furthermore, in order to boost our predictions, we proposed the use of two ensemble methods: the Majority Vote and the Highest Sum.

Based on \cite{magnossao-de-paula-etal-2022-upv, depaula2021exist, depaula2022exist}, we used a 0.00001 learning rate and a 0.3 dropout percentage for the transformer’s fine-tuning. We adopted a max length of 128 tokens and a batch size of 18 samples during all experiments. In order to find the suitable number of fine-tuning epochs, we carried out a five-fold cross-validation on the training data based on the task 1.d official metric, F1-score on the Hateful class. Table \ref{tab:epohcs} displays the optimal number of fine-tuning epochs for each transformer model.

\begin{table*}[!htp]
    \caption{Models’ best epochs}
    \label{tab:epohcs}
        \begin{tabular}{lc}
            \toprule
            \textbf{Model} & \textbf{Epochs} \\ \midrule
            AraBERT        & 4               \\
            AraELECTRA     & 3               \\
            Albert-Arabic  & 4               \\
            AraGPT2        & 4               \\
            mBERT          & 3               \\
            XLM-RoBERTa    & 1               \\
            \bottomrule
        \end{tabular}
\end{table*}

In Table \ref{tab:cross-validation}, we evaluate the best models during the cross-validation in terms of F1-score, Precision and Recall calculated for the Hateful class, and also the Accuracy, which considers the two classes (Hateful and Not Hateful). Additionally, as we previously mentioned, we implemented two ensemble methods. The Highest Sum method aggregates the transformers’ output values separately for each class and then selects the class with the highest sum. The Majority Vote method chooses the most predicted class among the transformers, and if there is a tie, it randomly selects one of the tied classes \cite{magnossao-de-paula-etal-2022-upv}.  

\begin{table*}[!htp]
    \caption{Training data experiment results using five-fold cross-validation}
    \label{tab:cross-validation}
        \begin{tabular}{llccccc}
        \toprule
        \multicolumn{2}{c}{\textbf{Models}}        &             & \textbf{F1-score} & \textbf{Acc.} & \textbf{Precision} & \textbf{Recall} \\
        \midrule 
        \multicolumn{1}{c}{\multirow{2}{*}{\textbf{Ensembles}}} & Majority Vote  &                     & \textbf{0.76}     & \textbf{0.95} & \textbf{0.88}      & 0.69            \\
        \multicolumn{1}{c}{}                                    & Highest Sum     &                    & 0.62              & 0.87          & 0.45               & 0.96            \\ \midrule
        \multirow{6}{*}{\textbf{Transformers}}                  & AraBERT          &                   & 0.68              & \textbf{0.95} & 0.68               & 0.68            \\
                                                                & AraGPT2           &                  & 0.61              & 0.93          & 0.80               & 0.50            \\
                                                                & AraELECTRA         &                 & 0.21              & 0.17          & 0.12               & 0.96            \\
                                                                & Arabic-ALBERT       &                & 0.20              & 0.11          & 0.11               & \textbf{1.00}   \\
                                                                & mBERT                &               & 0.20              & 0.11          & 0.11               & \textbf{1.00}   \\
                                                                & XLM-RoBERTa           &              & 0.20              & 0.11          & 0.11               & \textbf{1.00}   \\ \bottomrule
        \end{tabular}
\end{table*}

Analysing Table \ref{tab:cross-validation}, we can see that the transformer with the best performance regarding F1-score is AraBERT followed by AraGPT2 and AraELECTRA. The other transformers presented a similar performance. The two ensembles also presented competitive results, achieving the first (Majority Vote) and the third (Highest Sum) best F1-score. The Majority Vote ensemble presented impressive results as it achieved the highest Accuracy and Precision. On the other hand, AraELECTRA, Arabic-ALBERT, mBERT, and XLM-RoBERTa achieved good results when it comes to the Recall while performing poorly in the Accuracy, F1-score, and Precision. During fine-tuning, these models focused on predicting the positive class (Hateful expressions), while sacrificing most predictions for the negative class (Not Hateful). Due to the nature of the recall metric, only the samples that belong to the positive class, i.e., hateful samples, were considered. This explains why, despite obtaining high recall results, the rest of the metrics, which take into consideration the prediction of the negative class samples, presented lower performance scores.

The CERIST NLP Challenge 2022 accepts, for each participant in the task, only one submission of the test set predictions. Hence, we utilized the model with the best performance in Table 3 regarding F1-score, the Majority Vote ensemble. The organizers communicated that we achieved a 0.60 F1-score and a 0.86 Accuracy in the official test data, which aligns with our results in the training data.

\section{Conclusion}

This work addressed the problem of hate speech detection for Arabic language by applying six transformer models: AraBERT, AraELECTRA, Albert-Arabic,  AraGPT2, mBERT, and XLM-RoBERTa. We also took advantage of the Majority Vote and Highest Sum ensembles to aggregate the transformer's output and improve our final results. Based on the task 1.d official evaluation metric, AraBERT performed the best among the transformers, and Majority Vote ensemble achieved the highest score among all models using the five-fold cross-validation approach on the training data. Hence, we applied Majority Vote to carry out the official prediction based on the test data. In general, the Majority Vote ensembles presented a more robust performance in this task compared with the single transformers approach.

\begin{acknowledgments}
This publication was made possible by the NPRP grant 13S-0206-200281 (Resources and Applications for Detecting and Classifying Polarized and Hate Speech in Arabic Social Media) from the Qatar National Research Fund (a member of Qatar Foundation). The findings achieved herein are solely the responsibility of the authors.

\end{acknowledgments}

\bibliography{References_CERIST}

\appendix

\end{document}